\documentclass{IEEEtran4PSCC}
\usepackage{cite}
\usepackage{hyperref}       
\usepackage{url}            
\usepackage{booktabs}       
\usepackage{amsfonts}       
\usepackage{nicefrac}       
\usepackage{microtype}      
\usepackage{subcaption}
\usepackage{caption}
\usepackage{amsmath,mathrsfs,amssymb,mathtools,bm}
\usepackage{scalerel}
\usepackage{amsthm}
\usepackage{txfonts}
\usepackage{fontawesome}
\usepackage{wrapfig}
\usepackage{todonotes}
\usepackage{xcolor}
\usepackage{setspace}
\usepackage{bm,bbm}
\usepackage{paralist}
\usepackage{enumitem}

\usepackage[ruled,linesnumbered,noresetcount,vlined]{algorithm2e}
\makeatletter

\newcommand{\nosemic}{\renewcommand{\@endalgocfline}{\relax}}
\newcommand{\dosemic}{\renewcommand{\@endalgocfline}{\algocf@endline}}
\let\oldnl\nl
\newcommand{\nonl}{\renewcommand{\nl}{\let\nl\oldnl}}

\SetKwFor{Repeat}{repeat}{:}{endw}
\makeatother

\makeatletter
\newcommand{\oset}[3][0ex]{%
  \mathrel{\mathop{#3}\limits^{
    \vbox to#1{\kern-2\ex@
    \hbox{$\scriptstyle#2$}\vss}}}}
\makeatother

\usepackage{letltxmacro}
\LetLtxMacro\orgvdots\vdots
\LetLtxMacro\orgddots\ddots

\newtheorem{problem*}{Problem}

\newtheorem{theorem}{Theorem}

\newtheorem{definition}{Definition}

\newtheorem*{example*}{Example}

\DeclareMathOperator*{\argmin}{argmin}

\usepackage{multirow,mdframed}
\mdfsetup{skipabove=0pt,skipbelow=0pt}
\mdfdefinestyle{model}{%
  innertopmargin=0pt,
  innerbottommargin=0pt,
  innerleftmargin=0pt,
  innerrightmargin=0pt,
  skipbelow=0pt,%
  skipabove=0pt,
  splitbottomskip=0pt,
  splittopskip=0pt,
  leftmargin =0pt,%
    rightmargin=0pt,
  splittopskip=0pt,
    usetwoside=false,
}
\usepackage{float}
\floatstyle{ruled}
\newfloat{model}{thp}{lop}
\floatname{model}{Model}

\newcommand{\cC}{\mathcal{C}}

 \newcommand{\cL}{\mathcal{L}}
 
\newcommand{\cO}{\mathcal{O}}

 \newcommand{\RR}{\mathbb{R}}

\usepackage{esvect}

\makeatletter
\let\old@ps@headings\ps@headings
\let\old@ps@IEEEtitlepagestyle\ps@IEEEtitlepagestyle
\def\psccfooter#1{%
    \def\ps@headings{%
        \old@ps@headings%
        \def\@oddfoot{\strut\hfill#1\hfill\strut}%
        \def\@evenfoot{\strut\hfill#1\hfill\strut}%
    }%
    \def\ps@IEEEtitlepagestyle{%
        \old@ps@IEEEtitlepagestyle%
        \def\@oddfoot{\strut\hfill#1\hfill\strut}%
        \def\@evenfoot{\strut\hfill#1\hfill\strut}%
    }%
    \ps@headings%
}
\makeatother

\psccfooter{%
        \parbox{\textwidth}{\hrulefill \\ \small{Preprint. Under review.}}%
}

\begin{document}
%
\title{Towards Understanding the Unreasonable Effectiveness of Learning AC-OPF Solutions}

\author{
\IEEEauthorblockN{My H.~Dinh, Ferdinando Fioretto}
\IEEEauthorblockA{Syracuse University\\
Syracuse, NY, USA\\
\{mydinh, ffiorett\}@syr.edu}
\and
\IEEEauthorblockN{Mostafa Mohammadian, Kyri Baker}
\IEEEauthorblockA{University of Colorado Boulder\\
Boulder, CO, USA\\
\{mostafa.mohammadian, kyri.baker\}@colorado.edu}
}

\maketitle\sloppy\allowdisplaybreaks

\begin{abstract}

Optimal Power Flow (OPF) is a fundamental problem in power systems. 
It is computationally challenging and a recent line of research has proposed the use of Deep Neural Networks (DNNs) to find OPF approximations at vastly reduced runtimes, when compared to those obtained by classical optimization methods.
While these works show encouraging results in terms of accuracy and runtime, little is known on why these models can predict OPF solutions accurately, as well as about their robustness. This paper provides a step forward to address this knowledge gap. 
The paper connects the volatility of the generators outputs to the ability of a learning model to approximate them, it sheds light on the characteristics affecting the DNN models to learn good predictors, and it proposes a new model that exploits the observations made by this paper to produce accurate and robust OPF predictions.
\end{abstract}



\section{Introduction}
\label{sec:intro}

The Optimal Power Flow (OPF) problem finds the generator dispatch of minimal cost that meets the demands in a power system. The problem is required to satisfy the AC power flow equations, which are non-convex and nonlinear, 
and is a core building block in many power system applications. 
While its resolution has benefited from decades of 
research in power systems and operational research, the 
introduction of intermittent renewable energy sources 
is forcing system operators to adjust the generators 
set-points with increasing frequency. 
However, the resolution frequency to solve OPFs is 
limited by their computational complexity. To address this issue, system operators typically solve OPF approximations, such as the linear DC model, but, while more efficient computationally, their solutions may be sub-optimal and induce substantial economical losses.

Recently, an interesting line of research has focused on how to
approximate AC-OPF using Deep Neural Networks
(DNNs)~\cite{Deka19,Zamzam_learn_19,fioretto2020predicting}. Once a DNN is trained, predictions can be computed on the order of milliseconds.  
While the recent results show that these learning models can approximate the generator set-points of AC-OPF with high accuracy, little is known on why these models can predict OPF solutions accurately, as well as about their predictions robustness. 
This paper provides a step forward to address this knowledge gap and makes four main contributions.

\begin{figure}[t!]
\vspace*{-12pt}
\centering
\includegraphics[width=\columnwidth,height=120pt]{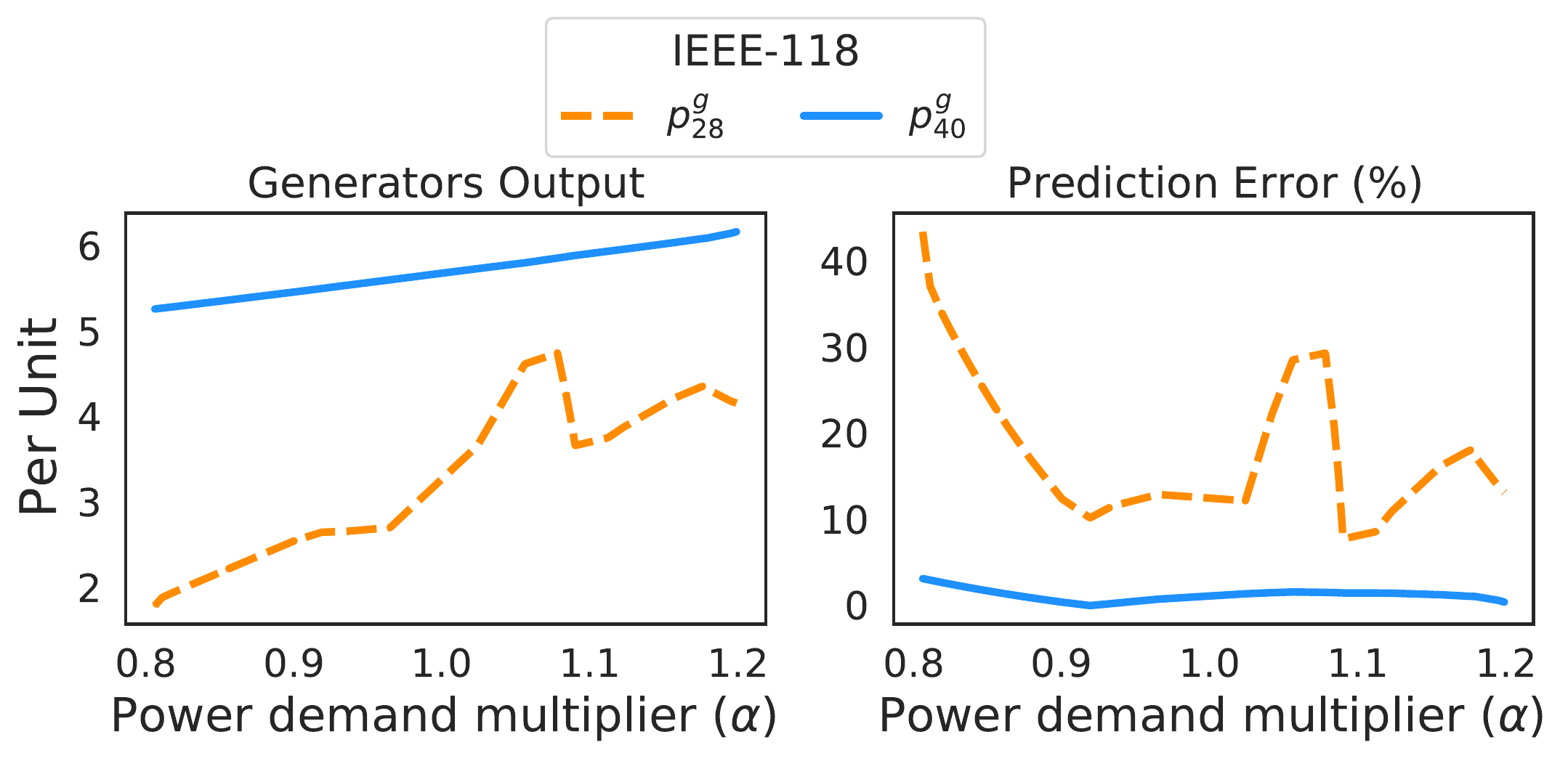}
\vspace*{-18pt}
\caption{Generator output as a function of demand (right) and associated predictions (left). Orange (blue) colors show high (low) volatile curves while continuous (dashed) lines depict easy (hard)  prediction tasks.}
\label{fig:motivation}\vspace{-10pt}
\end{figure}

\smallskip It firstly asks:
{\em Why are DNNs able to approximate OPF solutions with low errors?} 
To answer this question, the paper studies the relation between the training 
data and their target outputs. 
Figure~\ref{fig:motivation} (left) shows how generator outputs change as a function of the total demand for selected IEEE-118 generators. 
Notice that the blue curve suggests a linear dependence between the 
associated generator outputs and the loads, indicating that a simple learning model may effectively capture such behavior, as indeed confirmed in the corresponding low DNN prediction errors reported in Figure~\ref{fig:motivation} (right). 
{\em The paper shows that when many generators exhibit this behavior, approximating OPF with DNNs produces accurate results, on average}. 


\smallskip There are, however, also generators 
whose outputs are inherently more difficult to predict. The orange 
curve in the figure depicts a much different scenario with a more 
volatile underlying function. The right plot shows the high 
prediction error attained, indicating {robustness issues}. 
{\em The paper sheds light on why these behaviors are not easily 
captured by standard learning models connecting the stability of the training data to the ability of a learning model to approximate it.}

\smallskip 
Next, the paper asks: \emph{What are the latent factors that affect the prediction accuracy of these generators?} 
To address this question, the paper studies which characteristics of the OPF may be responsible for these erroneous predictions, and indicates the need for modeling and predicting the behavior of the OPF engineering and physical constraints during training to capture the complexity of the predictions. 

\smallskip Finally, in light of the robustness issues observed in this study, the paper proposes a new framework that relies on a deep autoregressive Recurrent Neural Network to exploit the data generated by iterative nonlinear solvers during training. 
The results show that this framework is not only able to improve the prediction robustness over existing DNN OPF predictors, but also it comes with a reduced memory footprint, thus, enabling it to predict very large instances,
 overcoming one of the limitations of existing DNN OPF predictors relying on fully-connected networks.

\section{Related Work}
\label{sec:related_work}

The use of machine learning to accelerate the resolution of power system optimization procedures has recently seen a growing number of results. A recent survey by Hasan et al.~\cite{hasan20survey} summarizes the  development in the area. 

In particular, Pan et al.~\cite{pan19deepopf} explore DNN architectures for predicting DC-OPFs, a linear approximation of the full AC model. Deka et al.~\cite{deka2019learning} and Ng et al.~\cite{ng2018statistical} use a DNN architecture to learn the set
of active constraints.  By exploiting the linearity of the DC-OPF
problem, once the set of relevant active constraints is identified,
an exhaustive search can be used to find a solution that satisfies the
active constraints.  A deep learning approach for AC-OPFs is also proposed by Yang et al.~\cite{8887248} to predict voltages and flows. This
approach focuses on specific operational constraints while dismissing other physical and engineering constraints.  

Other recent approaches have attempted to incorporate structure from OPF constraints into deep learning–based models. 
For instance, Fioretto et al.~\cite{fioretto2020predicting} propose 
a learning method which combines deep learning and Lagrangian duality, incorporating information about OPF dual variables into the learning loss function to promote the prediction of feasible solutions.
Other approaches focus on enforcing OPF constraints directly within the learning process. For instance, Zamzam and Baker \cite{Zamzam_learn_19} use a DNN to predict a partial OPF solution, and then solve for the remaining outputs using power the flow equations. 
Donti et al.~\cite{donti2020dc3} extended this approach though the use of implicit layers wdhich allows a DNN to reason about the hard constraints.

While these proposals have clearly shown that it is possible to approximate OPF solutions of high quality, and in vastly reduced computational times when compared to those required by traditional optimization solvers, a complete understanding of the reasons 
for the effectiveness of these learning models is missing. The rest of the paper provides a first step toward addressing this knowledge gap.

\section{Preliminaries}
\label{sec:opf_bk}

\noindent{\bf Optimal Power Flow}.
\emph{Optimal Power Flow (OPF)} is the problem of determining the least-cost 
generator dispatch that meets the demands in a power network.  A power
network is viewed as a graph $(N,E)$ where the set of nodes $n$ describes $n$ buses and the edges $E$ describe $e$ transmission lines. Here $E$ is a set of directed arcs and
$E^R$ is used to denote the arcs in $E$ but in reverse direction.

The AC power flow equations are based on complex quantities for
current $I$, voltage $V$, admittance $Y$, and power $S$. The
quantities are linked by constraints expressing Kirchhoff's Current
Law (KCL), i.e., $ I^g_i - {\bm I^d_i} = \sum_{\substack{(i,j)\in E
\cup E^R}} I_{ij},\label{eq:kcl}$ Ohm's Law, i.e., $ I_{ij} = \bm
Y_{ij} (V_i - V_j), \label{eq:ohm}$, and the definition of AC power,
i.e., $ S_{ij} = V_{i}I_{ij}^*.\label{eq:power}$
Combining these three properties yields the AC Power Flow equations, i.e.,
\begin{align*}
    & S^g_i - {\bm S^d_i} = \sum_{\substack{(i,j)\in E \cup E^R}} S_{ij} \;\; \forall i\in N \\ 
    & S_{ij} = \bm Y^*_{ij} |V_i|^2 - \bm Y^*_{ij} V_i V^*_j \;\; (i,j)\in E \cup E^R  
\end{align*}
%

\noindent These non-convex nonlinear equations are the core building
blocks in many power system applications. Practical applications
typically include various operational constraints on the flow of
power, which are captured in the AC OPF formulation in
Model~\ref{model:ac_opf}.  The objective function \eqref{ac_obj}
captures the cost of the generator dispatch.  
Constraints \eqref{eq:ac_1} and
\eqref{eq:ac_6} capture the voltage and phase angle difference
operational constraints.  Constraints \eqref{eq:ac_2} and
\eqref{eq:ac_5} enforce the generator output and line flow limits.
Finally, constraints \eqref{eq:ac_3} capture KCL and constraints
\eqref{eq:ac_4} capture Ohm's Law. 
\begin{model}[!t]
    \caption{The AC Optimal Power Flow Problem (AC-OPF)}
    \label{model:ac_opf}
    \vspace{-10pt}
    {\small
    \begin{align}
        \mbox{\bf variables:} \;\;
        & S^g_i, V_i \;\; \forall i\in N, \;\;
          S_{ij}     \;\; \forall(i,j)\in E \cup E^R \nonumber \\
        \mbox{\bf minimize:} \;\;
        & \sum_{i \in N} \bm c_{2i} (\Re(S^g_i))^2 + \bm c_{1i}\Re(S^g_i) + \bm c_{0i} \label{ac_obj} \\
        \mbox{\bf subject to:} \;\; 
        & \bm {v^l}_i \leq |V_i| \leq \bm {v^u}_i       \;\; \forall i \in N \label{eq:ac_1} \\
        & -\bm {\theta^\Delta}_{ij} \leq \angle (V_i V^*_j) \leq \bm {\theta^\Delta}_{ij} \;\; \forall (i,j) \in E  \label{eq:ac_6}  \\
        & \bm {S^{gl}}_i \leq S^g_i \leq \bm {S^{gu}}_i \;\; \forall i \in N \label{eq:ac_2}  \\
        & |S_{ij}| \leq \bm {s^u}_{ij}                  \;\; \forall (i,j) \in E \cup E^R \label{eq:ac_5}  \\
        & S^g_i - {\bm S^d_i} = \textstyle\sum_{(i,j)\in E \cup E^R} S_{ij} \;\; \forall i\in N \label{eq:ac_3}  \\ 
        & S_{ij} = \bm Y^*_{ij} |V_i|^2 - \bm Y^*_{ij} V_i V^*_j             \;\; \forall (i,j)\in E \cup E^R \label{eq:ac_4}
    \end{align}
    }
    \vspace{-14pt}
\end{model}
Notice that this is a non-convex nonlinear optimization problem 
and is NP-Hard \cite{verma2009power}.
Therefore, significant attention has been devoted to finding efficient
approximation of Model \ref{model:ac_opf}.

\smallskip\noindent{\bf Deep Learning Models}.
Supervised Deep Learning can be viewed as the task of
approximating a complex non-linear mapping from labeled data.  Deep
Neural Networks (DNNs) are deep learning architectures composed of a
sequence of layers, each typically taking as inputs the results of the
previous layer \cite{lecun2015deep}. Feed-forward neural networks are
basic DNNs where the layers are fully connected and the function
connecting the layer is given by
$
\bm{o} = \sigma(\bm{W} \bm{x} + \bm{b}),
$
where $\bm{x} \!\in\! \RR^n$ and is the input vector, $\bm{o} \!\in\! \RR^m$ the output vector, $\bm{W} \!\in\! \RR^{m \times n}$ a matrix of weights, and
$\bm{b} \!\in\! \RR^m$ a bias vector. The function $\sigma(\cdot)$ is often non-linear (e.g., a rectified linear unit (ReLU)).

\section{OPF Learning Goals}
\label{sec:opf_goals}

The goal of this paper is to analyze the effectiveness of learning an 
OPF mapping ${\cal O}: \RR^{2n} \to \RR^{2n}$: Given the loads
$\{\bm{S}_i^d\}_{i=1}^n$ (vectors of active and reactive power demand), predict the set-points $\{(\Re(S_i^g), |V_i|)\}_{i=1}^N$, of the generators, i.e., their active power and the voltage magnitude at their buses. 
In the following $\bm{p}^g$ and $\bm{v}$ are used as a shorthand for $\Re(S^g)$ and $|V|$.

The input of the learning task is a dataset ${\cal D} \!=\! \{(\bm{x}_\ell,\bm{y}_\ell)\}_{\ell\!=\!1}^N$, where
$\bm{x}_\ell \!=\! \bm{S}^d$ and $\bm{y}_\ell \!=\! (\bm{p}^g, \bm{v})$ represent the $\ell^{th}$ observation of load
demands and generator set-points which satisfy $\bm{y}_\ell \!=\! {\cal
O}(\bm{x}_\ell)$. The output is a function $\hat{\cal O}$ that ideally would be the result of the following constrained empirical minimization problem
\begin{subequations}
\begin{align}
\label{eq:ERPa}
\mbox{\bf minimize:} &\;\; \sum_{\ell=1}^N {\cal L}(\bm{y}_\ell,\hat{\cal O}(\bm{x}_\ell)) \\
\label{eq:ERPb}
\mbox{\bf subject to:} &\;\; {\cal C}(\bm{x}_\ell,\hat{\cal O}(\bm{x}_\ell)),
\end{align}
\end{subequations}
\noindent
where the loss function is specified by
\begin{equation*}
\label{basic_loss}
    {\cal L}(\bm{y}, \hat{\bm{y}}) = 
    {\| \bm{p}^g - \hat{\bm{p}}^g \|^2} +
    {\| \bm{v} - \hat{\bm{v}} \|^2},
\end{equation*}
and ${\cal C}(\bm{x},\hat{\bm{y}})$ holds if there exists voltage angles and reactive power generated that produce a
feasible solution to the OPF constraints with $\bm{x} =
\bm{S}^d$ and $\hat{\bm{y}} = (\hat{\bm{p}}^g, \hat{\bm{v}})$, where the \emph{hat} notation is adopted to denoted the predictions of the model.

One of the key difficulties of this learning task is the presence of
the complex nonlinear feasibility constraints in the OPF. The
approximation $\hat{\cal O}$ will typically focus on minimizing \eqref{eq:ERPa} while ignoring the OPF constraints or using penalty-based methods \cite{fioretto2020predicting}. Its predictions will thus not guarantee the satisfaction of the problem constraints. As a result, the validation of the learning task uses a 
load flow computation $\Pi_{C}$ that, given a prediction $\hat{\bm{y}}\!=\!\hat{\cal O}(\bm{x}_\ell)$, computes its projection onto the constraint set $C$, i.e., the closest feasible generator set-points $\Pi_{C}(\hat{\bm{y}}) = \argmin_{\bm{y} \in C} \|\hat{\bm{y}} - \bm{y}\|^2$, with $C$ being the OPF constraint set.

\section{Baseline Learning Model and Training Data}
\label{sec:baseline}

The baseline model for this paper assumes that the OPF approximation $\hat{\cal O}$ is given by a feed-forward fully connected (FCC) neural network, with $3$ hidden layers, each of size $4n$ and equipped with ReLU activations. This baseline model minimizes \eqref{eq:ERPa} but 
ignores the AC-OPF constraints $\cC(\bm{x}_\ell, \hat{\bm{y}}_\ell)$.
This baseline, as well as its variants described in Section \ref{sec:related_work}, often produce reliable and accurate predictions, albeit, as the paper will discuss in the next sections, not always robust. 

\noindent The next sections shed light on the reasons for these behaviors. Prior to do so, we describe the training data generation setting.

\smallskip\noindent{\textbf{Training Data}}
The paper analyzes the learning models behavior trained on test cases 
from the NESTA library \cite{Coffrin14Nesta}. For presentation simplicity, the analysis 
focuses primarily on the {IEEE 118, 162 and 300-bus networks}. 
However, the results are consistent across the entire benchmark set. 
The ground truth data are constructed as follows: For each network, 
different benchmarks are generated by altering the amount of nominal 
load $\bm{x} = \bm{S}^d$ within a range of $\pm 20\%$. 
For a given \emph{load multiplier} $\alpha$ sampled uniformly in 
the interval $[0.8, 1.2]$, a load vector $\bm{x}' = \bm{S}^{d'}$ is generated 
by perturbing each load value $\bm{S}^d_i$ independently with additive Gaussian noise centered in $\alpha$ and such that $\sum_i \bm{S}^{d'}_i = \alpha \sum_i \bm{S}^{d}_i$. 
A network value that constitute a dataset entry $(\bm{x}', \bm{y}' = 
\cO(\bm{x}'))$ is a feasible OPF solution obtained by solving the 
AC-OPF problem detailed in Model \ref{model:ac_opf}. 
The data are normalized using the per unit (pu) system. 
The experiments use a 80/20 train-test split and report results on 
the test set.

\section{Volatility Analysis of the Generators Dispatch}
\label{sec:sensitivity}


The first aspect being investigated concerns \emph{why deep learning models are able to approximate OPF solutions with low error}.
To answer this question, this section first analyzes the change 
in magnitude of the optimal generators dispatch at varying of the input loads and then relates this analysis to the complexity of learning to approximate the generators dispatch. Finally, the section will show that, for many test cases analyzed, the generators outputs exhibit low \emph{volatility}, enabling deep learning models to approximate them well. 

\smallskip
The following discussion assumes that the data point set 
$\{ \bm{x}_\ell\}_{\ell=1}^N$ is equipped with an ordering relation 
$\preceq$ such that $\bm{x} \preceq \bm{x}' \Rightarrow \|\bm{x}\|_p 
\leq \|\bm{x}'\|_p$ for some $p$-norm ($p \geq 1$).
Since the training data is generated by increasing or decreasing the 
network demand at each bus 
the ordering relation naturally applies to this domain.

Observe that, as illustrated in the motivating Figure~
\ref{fig:motivation}, the solution trajectory associated with the generator set-points on various input load parameters can often be naturally approximated by piecewise linear functions. 
The goal of the mapping function $\hat{\cO}$ is thus to approximate as best as possible these picewise linear functions associated with each generator's output. Intuitively, the more volatile the function is to approximate, the harder the associated learning task will be. This aspect will illustrated more formally in Section \ref{sec:theory}. To analyze this concept, the paper introduces the following notion.

\begin{definition}[Complexity Index]
Given a piecewise linear function $f:\mathbb{R}^k \to \mathbb{R}$ 
with $p$ pieces, each of width $h_i$ for $i \in [p]$, the complexity 
index (CI) of $f$ is a pair ${\sl CI}_f = (p, \omega)$, with $p$ being the number of its pieces and 
\[
\omega = \frac{1}{p} \sum_{i=1}^{p} h_i |L_{i} - L_{i-1}|,
\]
where $L_i$ is the slope of $f$ on piece $i$. Value $\omega$ describes the weighted average change in the slopes of $f$. 
\end{definition}

The complexity index allows us to reason about the \emph{volatility} of a piecewise linear function. It will become apparent later how this concept relates to the learning ability of ReLU neural networks.
Notice that the two piecewise linear functions can be compared, in terms of their volatility, by their associated complexity indexes using a lexicographic ordering. 

Since the generator dispatch trajectory can be approximated by a piecewise linear function, we refer to the complexity index of a generator $g$ to denote the complexity index of the induced piecewise linear function of the optimal dispatch $\cO(\bm{S}^d)$ of $g$ at varying of the loads $\bm{S}^d$ in the domain of interest. 

\begin{figure}[t!]
\centering
\includegraphics[width=0.9\columnwidth]{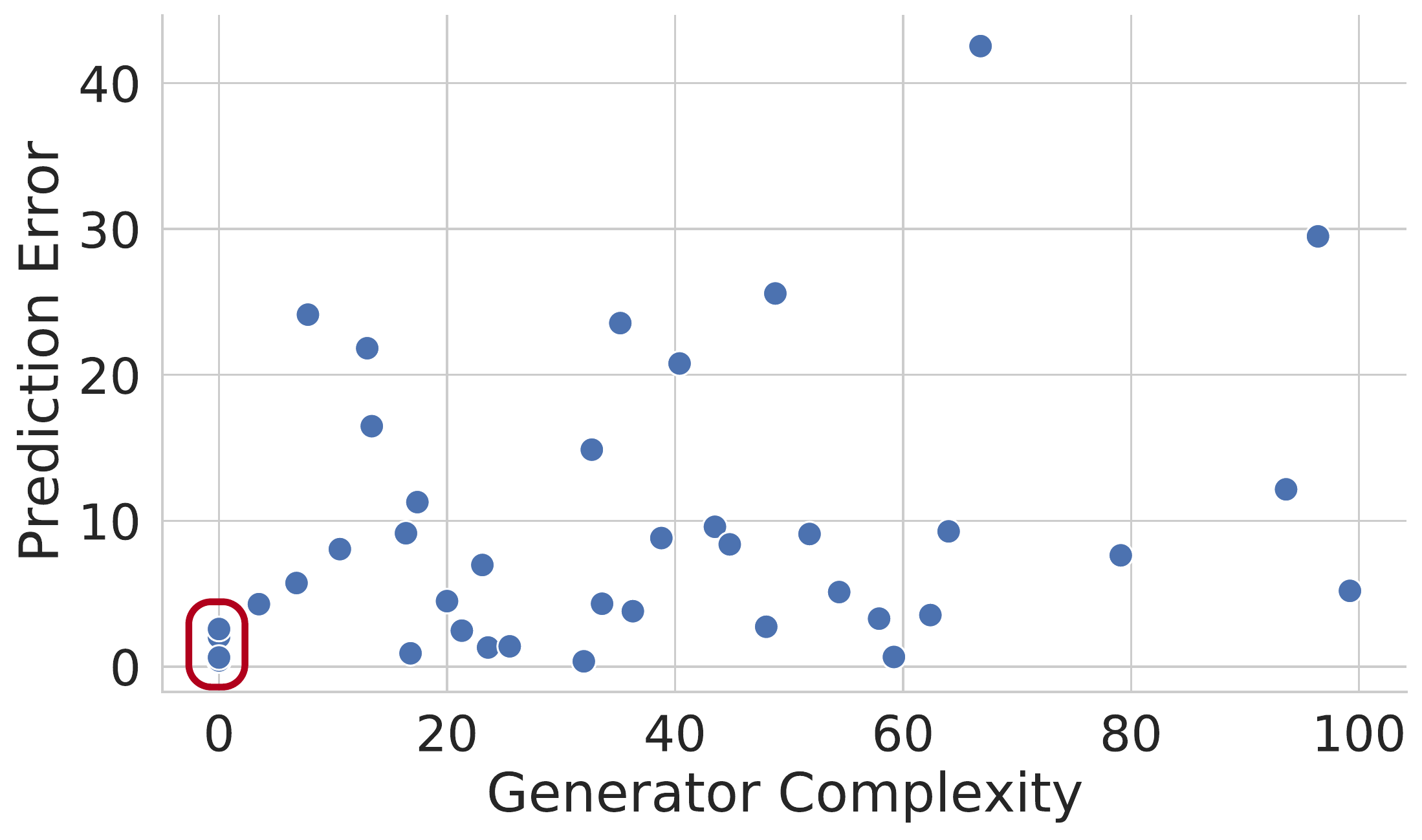}
\caption{Prediction errors, in percentage, of an FCC neural 
network. Generators are sorted by their CI values. The red box encloses generators with CI value $(1, \omega)$.}
\label{fig:gencomplexity}
\end{figure}

Figure \ref{fig:gencomplexity} illustrates the average prediction errors (in percentage) obtained when comparing the optimal dispatches $\bm{p}^g$, associated with different input load, to their predictions $\hat{\bm{p}}^g$ obtained by an FCC learning model as described in section \ref{sec:baseline}.
The figure reports the errors of each generators for test cases IEEE-118, -162, and -300 ordered by their (normalized) CI values.
Notice the strong correlation between the CI values and the model errors: \emph{More volatile generator dispatch trajectories correspond to generally less precise model predictions}.  
This observation connects the generators volatility with the hardness of the model to capture its output trajectory. 

In particular, notice that generators with a CI index of $(1, \omega)$ (enclosed in a red box in the figure) can be approximated by linear functions, and, thus, represent an ideal case for the learning task. 
The underlying models can be described using only two parameters (representing slope and intercept) and are generally characterized by low prediction errors.

This aspect is further emphasized in Table \ref{table:easy_gen}. 
The table reports the cumulative amount of generators whose trajectories are represented by a piecewise linear function with $1$, at most $2$, and at most $3$ pieces, the \emph{average prediction errors} $\|\hat{\bm{y}} - \bm{y}\|_1$ over the test set, 
the \emph{average load flow (LF) errors} $\|\Pi_{C}(\hat{\bm{y}}) - \bm{y}\|_1$ which compare the closest feasible solution $\Pi_{C}(\hat{\bm{y}})$ of the predictions $\hat{\bm{y}}$ with the optimal quantities $\bm{y}$, and the \emph{average optimality gap}, as
 $\frac{|O(\Pi_{C}(\hat{\bm{p}^g})) - O^*(\bm{p}^g)|}{O^*(\bm{p}^g)}$, with $O$ being the associated OPF cost.
First, note that many of the generators trajectories in the test cases analyzed can indeed be approximated by linear functions (i.e., their complexity index is $(1, \omega)$) or have CI with a low $p$ value (expressing the number of pieces of the associated piecewise linear function). 
Notice also that the predictions and load flow errors as well as the  optimality gap correlates positively with the amount of generators with larger complexity indexes. 

\emph{These observations shed light on why even simple fully connected ReLU networks, are able to approximate OPF solutions with relatively low average errors.}
The next section provides theoretical arguments to justify these observations.

\begin{table}[!t]
    \centering
    \resizebox{0.99\linewidth}{!}{
     \begin{tabular}{l | rrr rrr} 
     \toprule
     \multicolumn{1}{c}{\textbf{Test case}} & \multicolumn{3}{c}{CI p-value (\%)} 
     & Pred.~Err. & LF~Err. & Opt.~Gap\\
    \cmidrule(r){2-4}
    \multicolumn{1}{c}{} &  \multicolumn{1}{c}{$1$} & \multicolumn{1}{c}{$\leq 2$} & \multicolumn{1}{c}{$\leq3$} & (\%)& (\%)& (\%)\\ 
     \midrule
     IEEE-30 & 100.0 & 100.0& 100.0& 0.12& 0.128 &0.005\\ 
     IEEE-118 & 57.9 & 73.7 & 84.2 & 8.47 &   27.16 & 2.41\\
     IEEE-162 & 25.0 & 41.7 & 66.7 & 5.76 &  25.09 &2.06\\  
     IEEE-300 & 36.8 & 63.1 & 82.4 & 15.8 & 43.49  &6.23\\
     \bottomrule
    \end{tabular}  
}
\caption{CI and average errors of FCC model.}
\label{table:easy_gen}

\end{table}

\section{CI and Prediction Accuracy: Theoretical Insights} 
\label{sec:theory}
As observed above, the trajectory of the generators outputs can be described by piecewise linear functions. Next, note that ReLU networks capture piecewise linear functions~\cite{huang2020relu}.

\begin{figure}[tb]
\centering
\includegraphics[width = 0.9\linewidth]{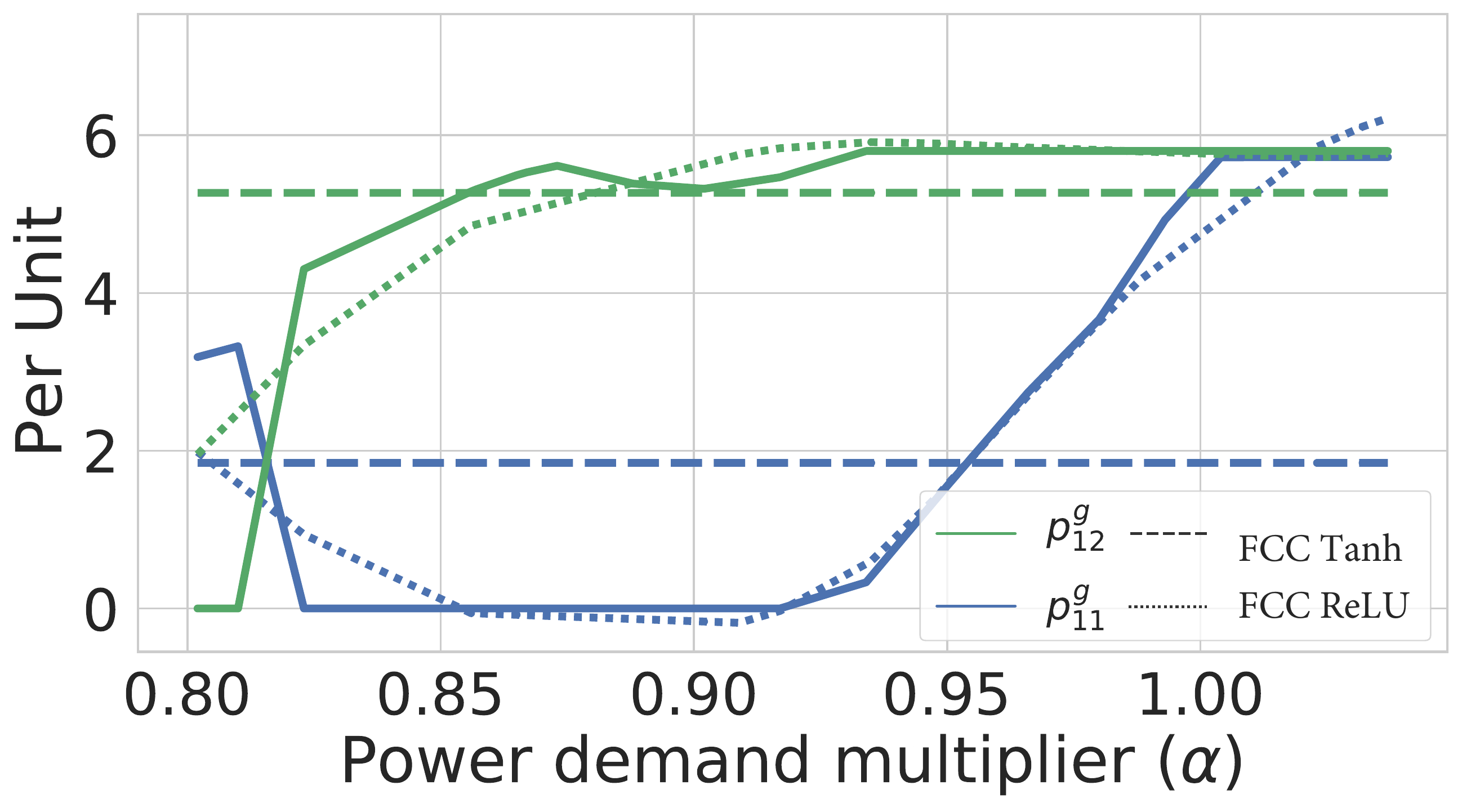}
\includegraphics[width = 0.9\linewidth]{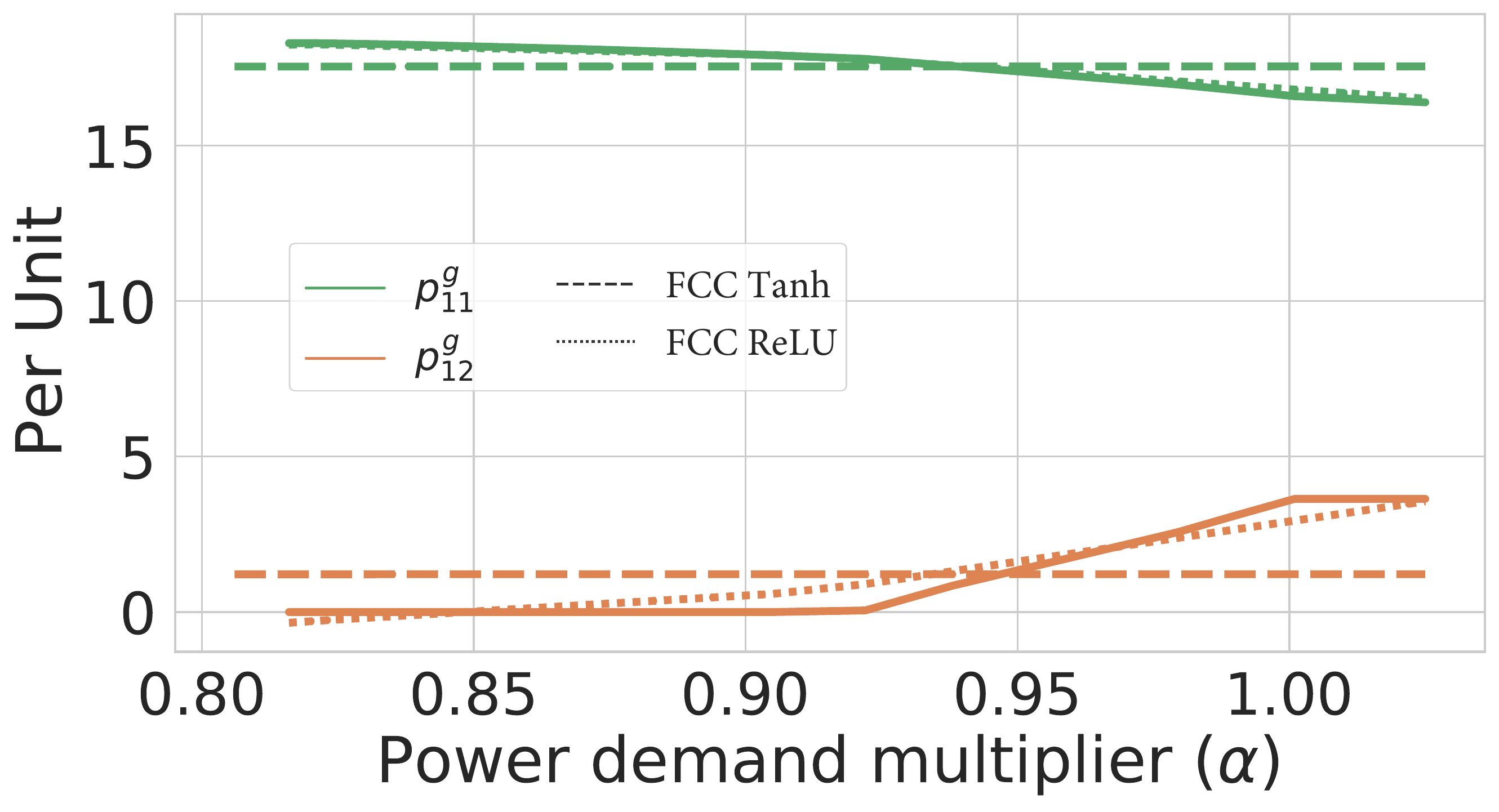}
\caption{Accuracy of ReLu FCC vs Tanh FCC on selected generators IEEE-162 (left) and IEEE-300 (right).}
\label{fig:activation}
\end{figure}

\emph{This observation justifies the choice of ReLU activation function for DNNs used to approximate OPF solutions}. Figure 
\ref{fig:activation} illustrates a comparison between two FCCs differing only in the type of activation functions they adopt. The plots show the original generators trajectories (solid lines), the approximations learned with a ReLU network (dotted lines) and those learned with a Tanh network (dashed lines). 
The top and bottom plots show results for selected generators from, respectively, the IEEE-162 and IEEE-300 test cases. Notice how the ReLU network predictions can represent piecewise linear functions that better approximate the original generator trajectories, when compared to those obtained from a Tanh network. 

While these ReLU FCC models are compatible with the task of predicting the solutions of an OPF problem, the model capacity required to represent a target piecewise linear function exactly depends directly on the number of constituent pieces.
Next, this section provides theoretical insights to link the ability of an FCC model to learn good approximations of generators trajectories of various CI complexities.

\begin{theorem}[Model Capacity \cite{arora2016understanding}]
Let $f : \RR^d \to \RR$ be a piecewise linear function with $p$ pieces. If $f$ is represented by a ReLU network with depth $k + 1$, then it must have size at least $\frac{1}{2}k p^{\frac{1}{k}}-1$. Conversely, any piecewise linear function $f$ that is represented by a ReLU network of depth $k + 1$ and size at most $s$, can have at
most $\left( \frac{2s}{k} \right)^k$ pieces.
\label{arora_lemma_D6}
\end{theorem}
The result above provides a lower bound on the model complexity to 
represent a given piecewise linear function. It implies that larger models may be able to better capture more complex relationships between inputs (loads) and output (generator set-points) values.

The second observation is from \cite{kotary:21b}. It relates the load values with the total variation of the generators outputs. The following theorem bounds the approximation error when using continuous piecewise linear functions: it connects the approximation errors of a piecewise linear function with the \emph{total variation in its slopes}. 

\begin{theorem}
\label{thm_james}
Suppose a piecewise linear function $f_{p'}$,  with $p'$ pieces each of width $h_k$ for $k \!\in\! [p']$, is used to approximate a piecewise linear $f_p$ with $p$ pieces, where $p' \!\leq\! p$. Then the approximation error 
$$\|f_p - f_{p'} \|_1 \leq \frac{1}{2}   h_{\max}^2  \sum_{1 \leq k \leq p} | L_{k+1} - L_k |,$$
holds where $L_k$ is the slope of $f_p$ on piece $k$ and $h_{\max}$ is the maximum width of all pieces. 
\end{theorem}  


The result above indicates that the more volatile the generators trajectory, the harder it will be to learn. Moreover, for a neural network of fixed size, the more volatile the generator trajectory, the larger the approximation error will be in general.

Combined with the observations reported in the previous section---showing that, for the test cases analyzed, a large number of generators have a low complexity index---the results above further illustrate the ability of DNNs to approximate OPF solutions with small average errors. 

\section{Robustness Issues}
\label{sec:challenges}

The results in the previous section are bounds on the ability of 
neural networks to represent generic functions. In practice, however, these bounds rarely guarantee the training of good approximators, 
as the ability to minimize the empirical risk (see Equation \eqref{eq:ERPa}) is often another significant source of error. 
This section demonstrates that there are also additional factors that may affect the ability of the DNN models to learn good OPF approximators, {including the presence of the OPF constraints}.

\begin{figure}[t]
    \centering
    \includegraphics[width = 0.9\linewidth]{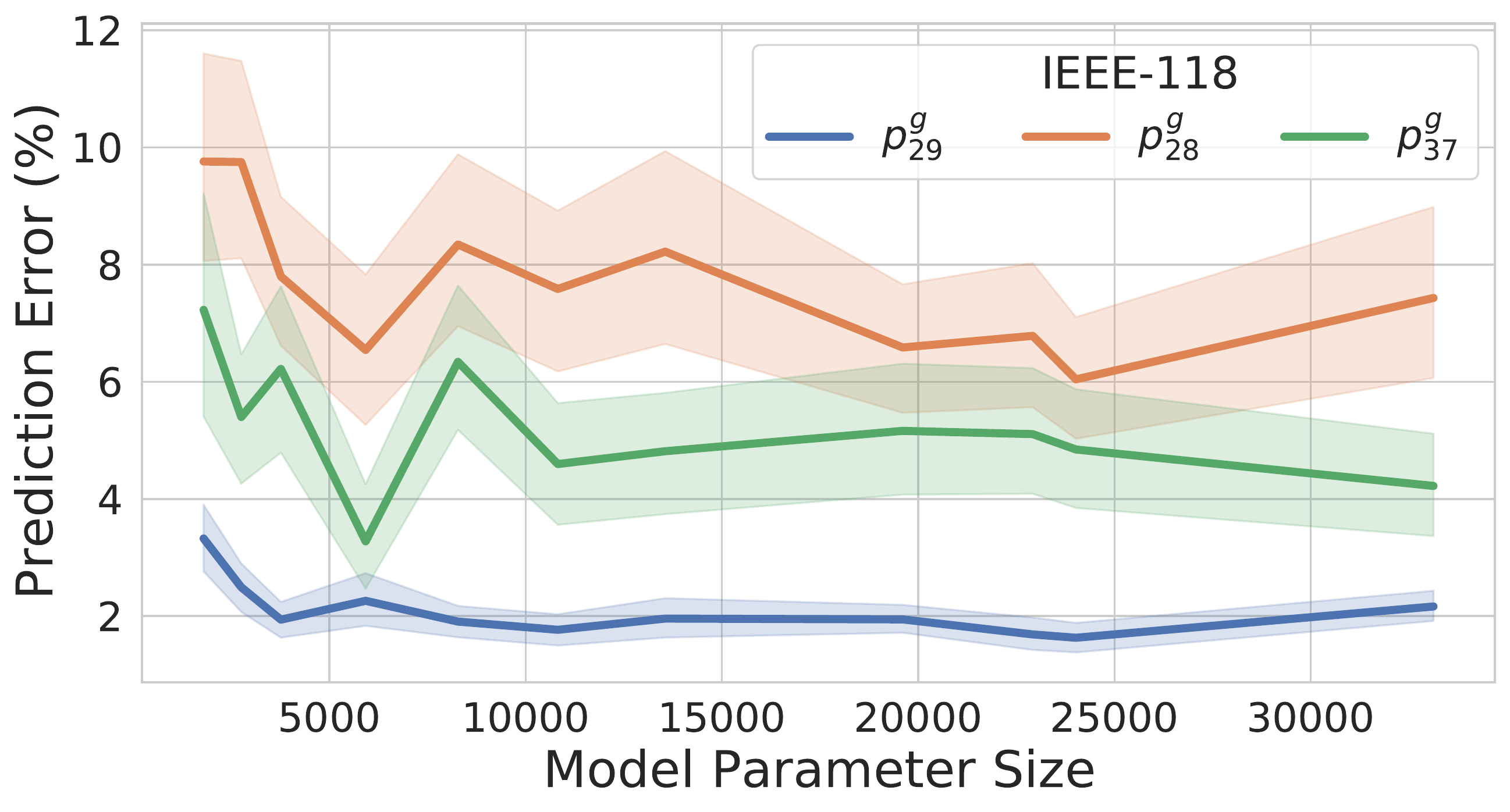}
    \caption{Prediction error for three key IEEE-118 generators at  increasing of the FCC model complexity.}
    \label{fig:model_size_acc}
\end{figure}

First notice that, in theory, 
it is to be expected that larger DNN models will be better suited to learning more complex solution trajectories 
(Theorem \ref{arora_lemma_D6}). However, this aspect was not observed in our experiments. Figure \ref{fig:model_size_acc} illustrates this surprising behavior. 
It reports the prediction errors associated with the trajectories of three IEEE-118 generators at the varying of the model size. 
Notice how prediction errors improvements saturate quickly and that even increasing the model size substantially does not produce notable error reductions. 
The reminder of the section seeks to answer \emph{why this behavior occurs}. 

\smallskip
To answer this question the paper analyzes generators with large complexity indexes. Indeed, high prediction errors pertain commonly to the solution trajectories associated with these generators.

\begin{figure}
\centering
\includegraphics[width = 0.9\linewidth]{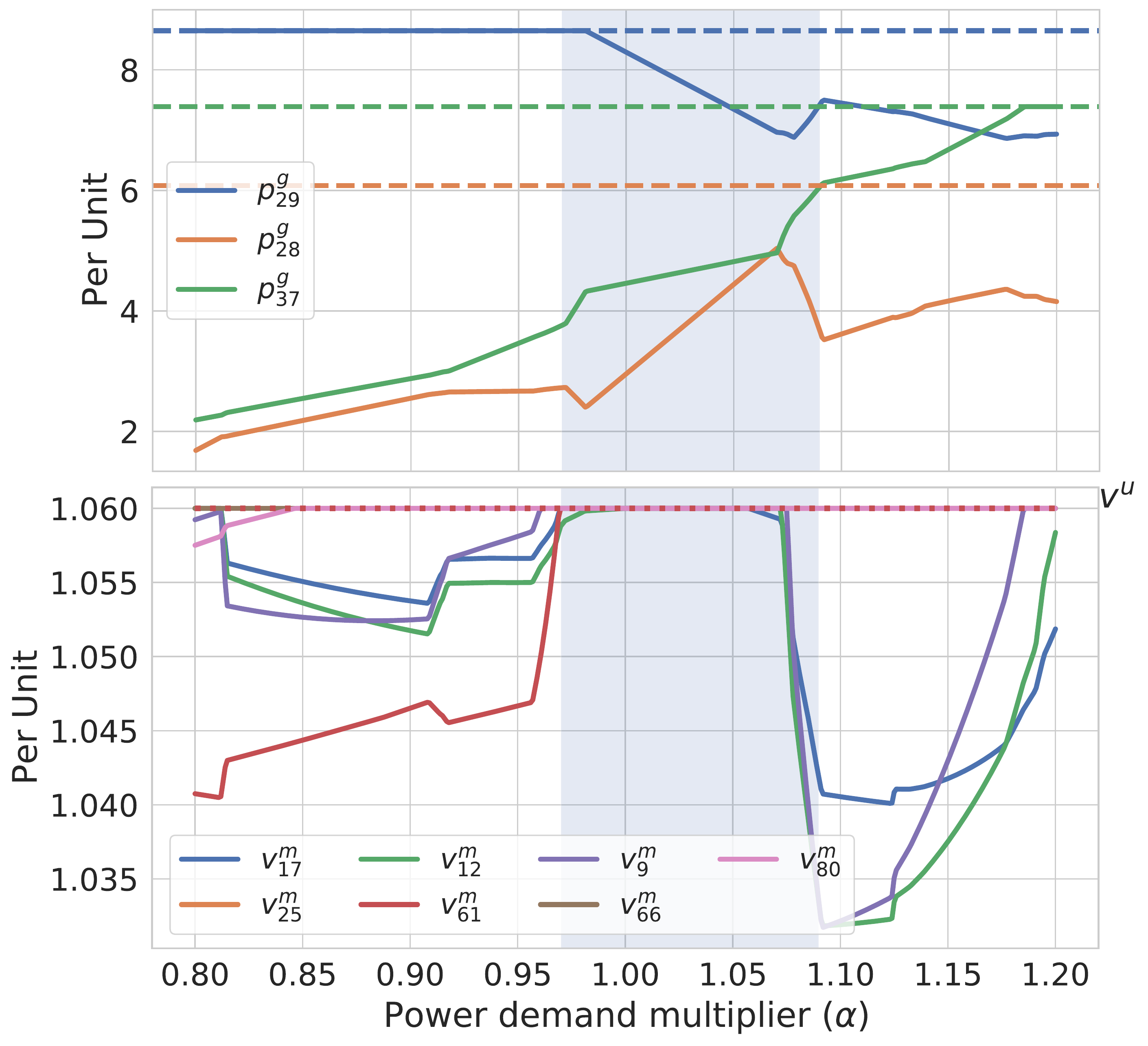}
\caption{Non-linear patterns of generators around load multiplier $\alpha \in [0.97, 1.05]$ (top) and associated voltage bounds issues 
at various buses.}
\label{fig:voltage_issue}
\end{figure}
Figure \ref{fig:voltage_issue} illustrates an example for the IEEE-118 test case, but these observations are consistent across the whole benchmark set analyzed. The figure highlights a region of \emph{high volatility} involving several generators. 
The top plot shows the dispatch trajectories of three generators (continuous lines) at varying of the input load multipliers $\alpha$ and their associated upper bound limits (dashed lines) (see constraints \eqref{eq:ac_2} of Model \ref{model:ac_opf}). The shaded area highlights the region in which large volatilities are observed. 
This region also correspond to the portion associated with the higher dispatch error predictions. 
The bottom plot shows the trajectories of the voltage magnitude values for a selection of buses. The upper bounds (constraints \eqref{eq:ac_1}) are illustrated with a dashed line. 
Notice that, while the generators dispatch are within the feasible operating regions, the bottom plot highlights the presence of voltage issues on several buses. The reported buses all are associated with voltage magnitudes value which results in binding constraints \eqref{eq:ac_1} in the region of high volatility of the generators considered.

\emph{These prediction errors are thus likely to arise as the hidden representation of the DNN does not accurately learn the operational and physical constraints which regulate the behavior of the OPF solutions}. In other words, the model is unaware of these constraints. 


Therefore, as investigated by several authors (including, \cite{Zamzam_learn_19,donti2020dc3,fioretto2020predicting}) this work found that actively exploiting the problem constraints during training to be an effective mechanism to enhance the model accuracy. 
The constraints were added using a model similar to \cite{fioretto2020predicting} which encourages the satisfaction of the OPF constraints by the means of a Lagrangian dual approach. 
Notice that the \emph{constrained} and baseline models differ solely in the loss function, and not in the number of their parameters. 

Table \ref{table:constraint} summarizes the results. It compares the average absolute constraint violations (in p.u.) for the set-points bounds (constraints \eqref{eq:ac_1} and \eqref{eq:ac_2}) and the KCL (constraint \eqref{eq:ac_3}), the load flow (LF) distance of the predictions $\hat{\bm{y}}$ to their optimal dispatches $\bm{y}$, and 
the optimality gaps, as defined in Section \ref{sec:sensitivity}.
Notice how the constrained model reduces the constraint violations, when compared to the baseline, as well as increases the associated prediction accuracy. 

\begin{table*}[!t]
\centering
\resizebox{\linewidth}{!}{
\begin{tabular}{l | rrrr | rrrr | rrrr}
    \toprule
        \multicolumn{1}{c}{\textbf{Test case}}     & 
        \multicolumn{4}{c}{FCC Without Constraint} &
        \multicolumn{4}{c}{FCC With Constraint}  &
        \multicolumn{4}{c}{RNN With Constraint}\\
    \cmidrule{2-5}\cmidrule{6-9}\cmidrule{10-13}
    & {Bound Vio} & {KLC Vio } & {LF Err.~(\%)}&{Opt. Gap (\%)}
    & {Bound Vio} & {KLC Vio } & {LF Err.~(\%)}& {Opt. Gap (\%)}
    & {Bound Vio} & {KLC Vio } & {LF Err.~(\%)}& {Opt. Gap (\%)}\\
    \midrule
     IEEE-30  & 0.000 & 0.001 & 0.128 & {0.005} & 0.000  & 0.081 & 0.080    & 0.001 & 0.0 & 0.13 & 0.384& 0.270\\ 
     IEEE-118 & 0.007 & 0.087 & 23.59 & 2.41    & 0.003  & 7.16  & 14.34    & 4.910 & 0.002 & 0.052 & 2.901 & 0.131\\  
     IEEE-162 & 0.047 & 0.363 & 25.83 & 2.06    & 0.016  & 8.35  & 19.17    & 2.191 & 0.012 & 0.038 & 4.478 &0.167 \\ 
     IEEE-300 & 0.000 & 0.015 & 0.205 & 17.34   & 6.23   & 0.021 & 11.56    & 16.89 & 0.023 & 0.0003 & 1.099 & 0.327\\ 
    \bottomrule
    \end{tabular} 
}
\caption{Accuracy comparison: FCC with and without constraints and RNN models.}
\label{table:constraint}
\end{table*}

\begin{figure}
\centering
\includegraphics[width = 0.9\linewidth]{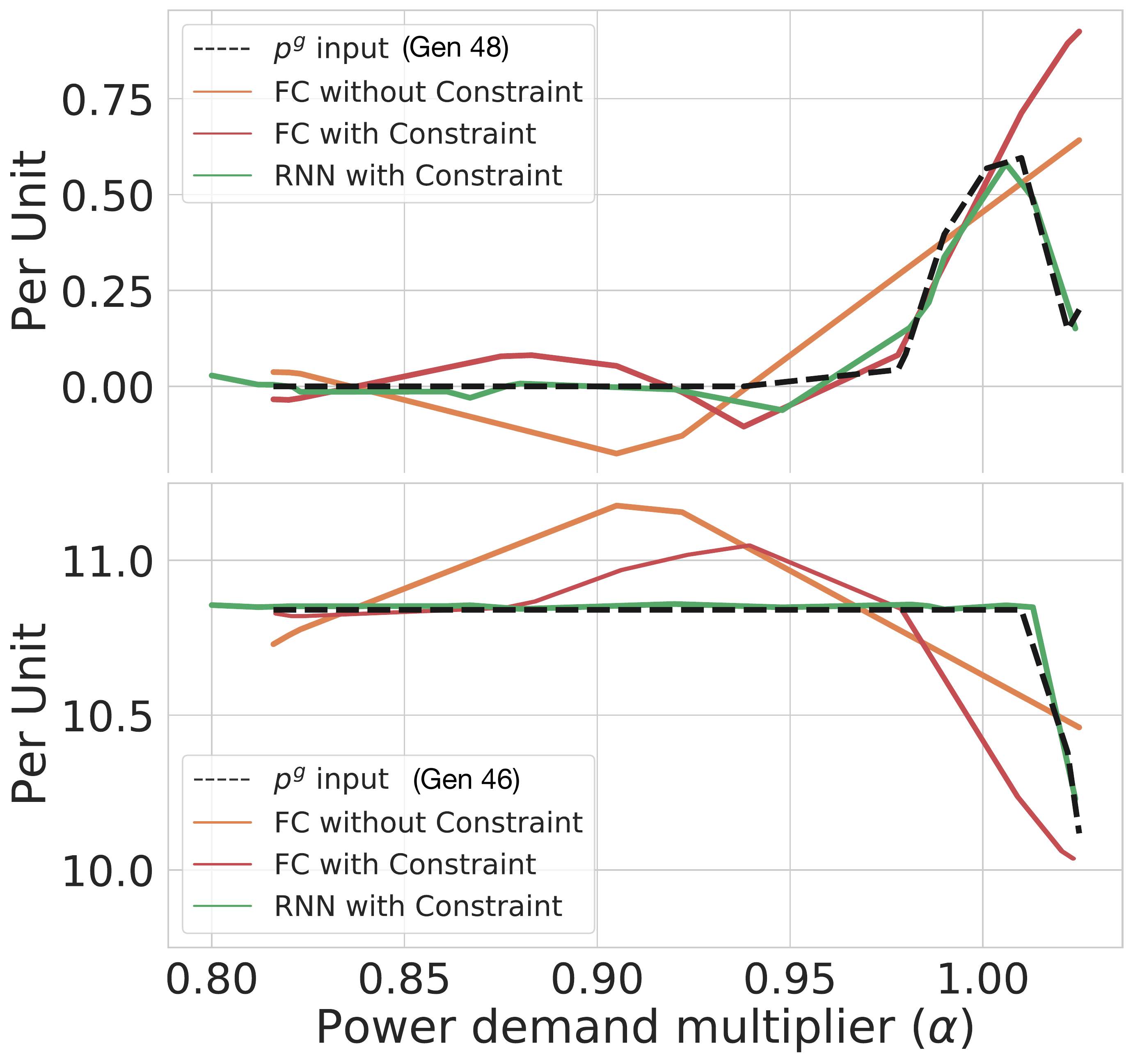}
\caption{IEEE-300. Optimal generators trajectory (red) for generator 36 (top) and 48 (bottom). Predictions: FCC without constraints (orange), FCC with constraints green), and RNN with constraint (blue). }
\label{fig:prediction_issue}
\end{figure}

This aspect is also evident in Figure \ref{fig:prediction_issue}, which compares the prediction trajectories of the FCC model with (yellow curves) and without (red curves) constraints for two high-complexity IEEE-300 generators. 
Notice that the constrained model predictions follow more closely the original trajectories when compared to the simple model. 

This aspect is surprising from an empirical risk minimization 
perspective: Including constraints using Lagrangian-based penalties adds additional terms to the loss function which can be interpreted as further regularizing terms, and thus, it may be expected they would reduce the model variance further. 

However, Figure \ref{fig:prediction_issue} also highlights some drawbacks of the constrained model. Despite its improved accuracy (and its ability to approximate precisely many \emph{easy} generators) its predictions tend to discard the rapid changes in trajectories of the generators outputs (see bottom plot). 
From a data representation point of view, these cases (where the change in trajectory occurs) represent \emph{outliers} and thus are  hard to predict. 
This observation motivates the introduction of a novel model described next.

\section{A Novel RNN-based Learning Framework}

The issue observed above could be partially addressed by providing additional training data to the learning task with the goal of more suitably representing the inputs associated with the \emph{outlier} set-points. Creating this data is, however, a very challenging task. It is unknown a-priori which set-point, within a trajectory, may be uncommon. Additionally, generating the input loads associated to a desired set-point would be an extremely challenging operation.

While generating additional targeted data is thus unfeasible, this section notices that iterative solvers, typically adopted to solve non-linear programs, generate a solution at each iteration of their execution. For example, IPOPT \cite{wachter06on}, a popular nonlinear solver adopted in this paper to generate OPF solutions, implements a primal-dual interior point line search filter method to find a local optimal solution to a given problem instance. 
The underlying idea of the proposed model is thus to exploit these \emph{solution trajectories} during training. 

To do so, this section introduces a DNN model for OPF predictions which relies on deep autoregressive Recurrent Neural Networks (RNN). 
RNNs are a powerful tool to learn from sequential data and have been vastly adopted in domains including natural language processing and computer vision \cite{10.1115/1.2847757, 279188, LIU2020113082}. 
An autoregressive model is typically used in time-series modeling where the current time step value $z_t$ depends linearly on some value $z_{t'}$ with $t' < t$. Similarly, autoregressive RNNs condition the prediction of the current time step on the predictions of the previous steps. 
They thus are a natural fit for the intended purpose.  

\begin{figure}
    \centering
    \includegraphics[width=\linewidth]{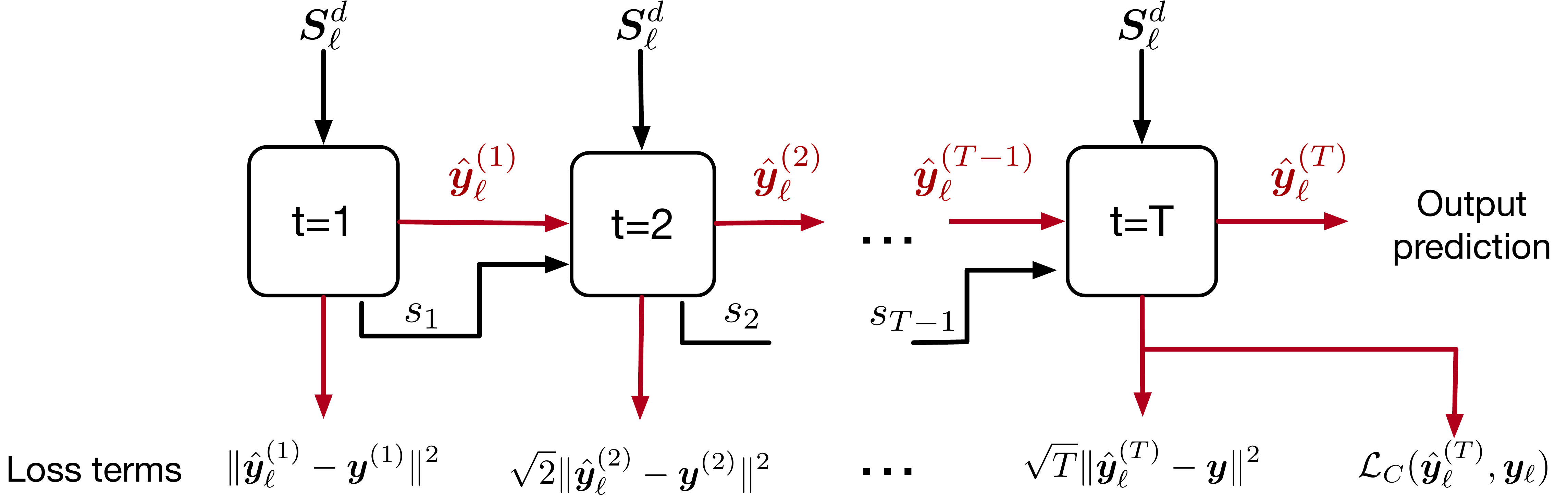}
    \caption{RNN model Overview.}
    \label{fig:rnn_model}
\end{figure}
The proposed model is illustrated in Figure \ref{fig:rnn_model}. 
The model is composed by $T$ sequential Long Short Memory Term (LSMT) units. For unit $t \in [T]$, the model takes as input the demands $\bm{x} = \bm{S}^d$ as well as 
the embedding $\bm{y}^{(t-1)}$ outputted by unit $t-1$, and the \emph{state} $s_{t-1}$ of unit $t-1$.
The first unit $t=1$ is special and only considers the input demands $\bm{x}$. 
The model uses the following loss:
\begin{equation}
\sum_{\ell=1}^N \sum_{t=1}^T
 \sqrt{t} \cL(\bm{y}^{(t)}_\ell, \hat{\bm{y}}^{(t)}_\ell) 
   + \cL_C(\bm{y}_\ell, \hat{\bm{y}}^{(T)}),
\end{equation}
where $\cL_C$ is the Lagrangian loss involving the prediction from the last unit to encourage constraint satisfaction, equivalently to that adopted by the constrained variant of the FCC model. The $\sqrt{t}$ multiplicative factor is adopted to give larger weights to the latter units. The model returns $\hat{\bm{y}}^{(T)}$ as its prediction, which is the output of the recurrent final unit. 

The predictions of the proposed model are summarized in Table \ref{table:constraint} (right). Notice how the model can reduce the load flow errors and optimality gaps by one order of magnitude when compared with the best FCC results. Notably, the RNN model predictions are much closer to satisfy the KLC than those produced by the constrained version of the FCC model. This is important as KLC are notoriously hard to satisfy for the predictions of DNN models \cite{fioretto2020predicting}.
The ability of this model to capture robustly rare changes in generators trajectory can be appreciated in Figure \ref{fig:prediction_issue}. 

\begin{table}[!t]
\centering
\resizebox{0.95\linewidth}{!}{
\begin{tabular}{l r r | l r r} 
     \toprule
     \textbf{Test Case} & FCC & RNN &
     \textbf{Test Case} & FCC & RNN \\ 
     \midrule
     IEEE-118 & 11.4  & 0.007 & IEEE-300 & 47.8 & 0.04    \\ 
     IEEE-162 & 14.8 &  0.005 & PEGASE-1354 & 154 & 0.32  \\ 
     EDIN-189 & 3.4 &  0.013  & RTE-2868 & 2907 & 1.64\\  
    \bottomrule
    \end{tabular}  
}
\caption{RNN vs FCC: Model parameter size (Million).}
\label{table:model_param}
\vspace{-10pt}
\end{table}

Finally, Table \ref{table:model_param} reports a comparison of the number of parameters (proxy to memory footprint) required by the FCC and the proposed RNN models. Notice that the FCC grow very large with the size of the processed test case highlighting scalability issues, as also observed in \cite{chatzos2020highfidelity}, which reported the inability of these models to fit in memory for test cases larger than $2000$ buses. 
In contrast, the proposed RNN model does not incur this drawback 
rendering it applicable to very large power systems. 


\section{Conclusions}
This paper was motivated by the recent development around using deep neural networks (DNN) to approximate the solutions of Optimal Power Flow (OPF) problems. While these learning models show encouraging results, little is known on why they predict OPF solutions accurately, as well as about their predictions robustness. 
The paper provided a step forward to address this knowledge gap.
It studied the connection between the volatility of the generators outputs with the ability of a learning model to approximate it, showing that many test cases are characterized by a large number of generators which are easy to predict. 
It also showed that operational and physical constraints are necessary to capture the complexity of the predictions. 
Finally, it proposed a new learning model based on recurrent neural networks, that was not only able improve the prediction accuracy over existing supervised learning approaches, but also reduced the memory requirements. 

\vspace{-6pt}
\section*{Acknowledgement}
This research is partially supported by NSF grant 2007164 and NSF CAREER award 2041835. 

\bibliographystyle{IEEEtran}
\bibliography{references}

\end{document}